%% file: acl_latex.tex
\pdfoutput=1

\documentclass[11pt,usenames,dvipsnames]{article}

\usepackage{acl}
\usepackage{times}
\usepackage{latexsym}
\usepackage{multirow}
\usepackage{hyperref}
\usepackage{times}
\usepackage{latexsym}
\usepackage[utf8]{inputenc}
\usepackage{xcolor,colortbl}
\usepackage{microtype}
\usepackage{url}
\usepackage{longtable}
\usepackage{tikz}
\usepackage{tabularx}
\usepackage{lscape}
\usepackage{hhline}
\usepackage{bbm}
\usepackage{array}
\usepackage{graphicx}
\usepackage{booktabs}
\usepackage{makecell}
\usepackage{amsfonts}       
\usepackage{nicefrac}       
\usepackage{microtype}      
\usepackage{multirow}
\usepackage{caption}
\usepackage{amsmath}
\usepackage{float} 
\usepackage{ragged2e}
\usepackage{breqn}
\usepackage{soul}
\usepackage{tabularx}
\usepackage{graphicx,subcaption} 
\usepackage[T1]{fontenc}

\usepackage[utf8]{inputenc}
\definecolor{pinkbox}{RGB}{255,182,193}
\definecolor{bluebox}{RGB}{100,149,237}
\definecolor{greenbox}{RGB}{0,100,0}
\definecolor{greybox}{RGB}{128, 128, 128}
\usepackage{microtype}

\title{IndicMT Eval: A Dataset to Meta-Evaluate \\
Machine Translation Metrics for Indian Languages}

\author{\textbf{Ananya B. Sai}$^1$ \quad
        \textbf{Tanay Dixit}$^1$ \quad
        \textbf{Vignesh Nagarajan}$^2$ \quad
        \textbf{Anoop Kunchukuttan}$^{2,4}$ \quad \\
        \textbf{Pratyush Kumar}$^{2,4}$ \quad \textbf{Mitesh M. Khapra}$^{1,2}$   \quad \textbf{Raj Dabre}$^3$ \quad \\
        Indian Institute of Technology Madras$^{1}$ \quad AI4Bharat$^{2}$ \\ 
        \quad National Institute of Information and Communications Technology$^{3}$ \quad Microsoft$^{4}$ \\
        \\
        {\tt ananya@cse.iitm.ac.in } \quad
        {\tt dixittanay@gmail.com}\\
        {\tt vignesh.vn.nagarajan@gmail.com
        } \quad
        {\tt ankunchu@microsoft.com
        }\\ 
        {\tt pratykumar@microsoft.com} \quad
        {\tt miteshk@cse.iitm.ac.in}
         \\
         {\tt raj.dabre@nict.go.jp}
} 

\begin{document}
\maketitle
\begin{abstract}
The rapid growth of machine translation (MT) systems necessitates meta-evaluations of evaluation metrics to enable selection of those that best reflect MT quality. Unfortunately, most meta-evaluation studies focus on European languages, the observations for which may not always apply to other languages. Indian languages, having over a billion speakers, are linguistically different from them, and to date, there are no such systematic studies focused solely on English to Indian language MT. This paper fills this gap through a Multidimensional Quality Metric (MQM) dataset consisting of 7000 fine-grained annotations, spanning 5 Indian languages and 7 MT systems. We evaluate 16 metrics and show that, pre-trained metrics like COMET have the highest correlations with annotator scores as opposed to n-gram metrics like BLEU.
We further leverage our MQM annotations to develop an Indic-COMET metric and show that it outperforms COMET counterparts in both human scores correlations and robustness scores in Indian languages. Additionally, we show that the Indic-COMET can outperform COMET on some unseen Indian languages. We hope that our dataset and analysis will facilitate further research in Indic MT evaluation.

\end{abstract}

\section{Introduction}
Natural language generation (NLG) has seen rapid progress in the past few years due to advancements in the field of large language models (LLMs) \cite{lewis-etal-2020-bart, liu-etal-2020-multilingual-denoising, dabre-etal-2022-indicbart, scao2022bloom}. Although initial research had focused on high-resource languages, recently the focus has shifted to middle-resource and low-resource languages. In the context of machine translation (MT), there is increasing interest in building massively multilingual models supporting numerous translation directions. For example, \citet{costa2022no} release a model which supports around 200 languages (40K directions). While this is commendable, to make MT truly inclusive, it is important that various design choices in the MT life-cycle are evaluated for low-resource languages and not simply transferred and adapted from English. One such important choice is of the correct evaluation metric to be used for evaluating MT systems. 



A recent survey by \citet{sai2022survey} has shown that over the last decade, many evaluation metrics have been proposed for MT. In parallel, several works \cite{callison-burch-etal-2006-evaluating_bleu_in_mt, sai-etal-2021-perturbation, mathur-etal-2020-tangled, tan-etal-2015-awkward, fabbri-etal-2021-summeval} have shown the inadequacy of popular metrics, such as BLEU \cite{bleu}, ROUGE \cite{rouge}. However, many languages are not represented in these works and most of the focus is on European languages. On the other hand, there is a growing body of work on machine translation focused on language groups such as Indian \cite{dabre-etal-2022-indicbart}, Indonesian \cite{cahyawijaya-etal-2021-indonlg}, and African \cite{reid-etal-2021-afromt}. However, these works rely on English centric metrics
due to lack of sufficient studies with tried and tested recommendations for their evaluation of  the languages. While techniques like MQM (Multidimensional Quality Metric) are being used for collecting better quality human-evaluation data for English and a select few other languages \cite{freitag-etal-2021-experts}, such multidimensional evaluations and analyses are not available for several language groups.

We narrow our focus to evaluation of one of these language groups, namely Indian languages which have more than a billion speakers worldwide. Indian languages are morphologically rich, especially Dravidian languages, which exhibit agglutination. Furthermore, they have relatively free-word order \cite{murthy-etal-2019-addressing, kunchukuttan2020utilizing} as compared to European languages which means that frequently used metrics such as BLEU may not always be reliable. This calls for an independent focused study on the evaluation of metrics for Indic languages in order to understand whether these conclusions drawn hold true for the Indian languages. 

In this paper, we aim to bridge this gap by focusing on the evaluation of MT (from English) into 5 Indian languages from 2 different families and make significant contributions towards designing MT evaluation metrics for these languages. Our main contribution is in the form of the MQM dataset for Indian languages created by taking outputs generated by 7 popular MT systems and asking human annotators to judge the quality of the translations using the MQM style guidelines \cite{mqm_lommel2014multidimensional}. With the help of language experts who are experienced in translation, we generate an MQM dataset consisting of 7000 annotated sentences, 1400 per language. 

We use the aforementioned dataset to establish correlations between the annotator scores and existing automatic metrics scores belonging to the following classes: (i)  n-gram and character based such as BLEU, METEOR, chrF++, (ii) embeddings based such as Vector Extrema, BERTScore, (iii) pre-trained metrics like BLEURT-20, COMET. We observe that pre-trained metrics show the highest correlations with the annotator scores, with the COMET metric performing the best (\S\ref{sec:results_1}). 
Additionally, we also observe that the metrics are not capable of capturing the fluency-based errors for Indian languages (\S\ref{sec:corr_flu_ade}). Finally, we use our data to train an Indic-COMET metric which not only shows stronger correlations with human judgement on Indian languages, but is also more robust to perturbations (\S\ref{sec:indic_comet}). We hope that our dataset and metric, which are publicly available\footnote{https://github.com/AI4Bharat/IndicMT-Eval}, will help spur research in this field. 

\input{s2RelatedWork}
\input{s3IndicEvalDataset}

\input{s4ExperimentalSetup}

\section{Results and Discussions}
In this section, we present the segment-level correlations in \S\ref{sec:results_1} and system-level correlations in \S\ref{sec:ranking_metrics}, followed by analyzing metrics in \S\ref{sec:metrics_spread}, \S\ref{sec:corr_flu_ade}.

\subsection{Segment-level Evaluation}
\label{sec:results_1}

\input{tab_sys_level_corrs}

The correlation between MQM-based scores and metric scores, measured using Pearson and Kendall-tau correlations on 1400 segments per language as shown in Table \ref{tab:corr_formula_metrics}. We observe that out of the overlap-based metrics, chrF++ has the highest correlation across all languages, but overall overlap-based metrics are the worst performing which is in line with the findings of \citet{kocmi-EtAl:2022:WMT}. Among the embedding-based metrics, LabSE embeddings yields better correlations than any of the other embedding-based approaches. The correlations improve further when we use BERTscore with embeddings obtained from different multilingual models. The results in this case are mixed, with MuRIL showing the best correlations on average. Overall, we observe that neural-network-based, end-to-end trained metrics with exposure to Indian languages are the best-performing metrics on average. The trained metric PRISM, which has been trained on 39 languages, out of which the only Indian language is Bengali, performs very poorly on all the 5 Indian languages in our study, partially owing to the minimal Bengali data used for training. On the other hand, BLEURT-20, a metric
finetuned on ratings from the WMT Metrics Shared Task and synthetic data from the WMT corpus, has fairly good correlations on all languages except Hindi. COMET-metric variants have the highest overall correlations for all the languages. 



\begin{figure}[h!]
    \centering
    \includegraphics[width=1\columnwidth]{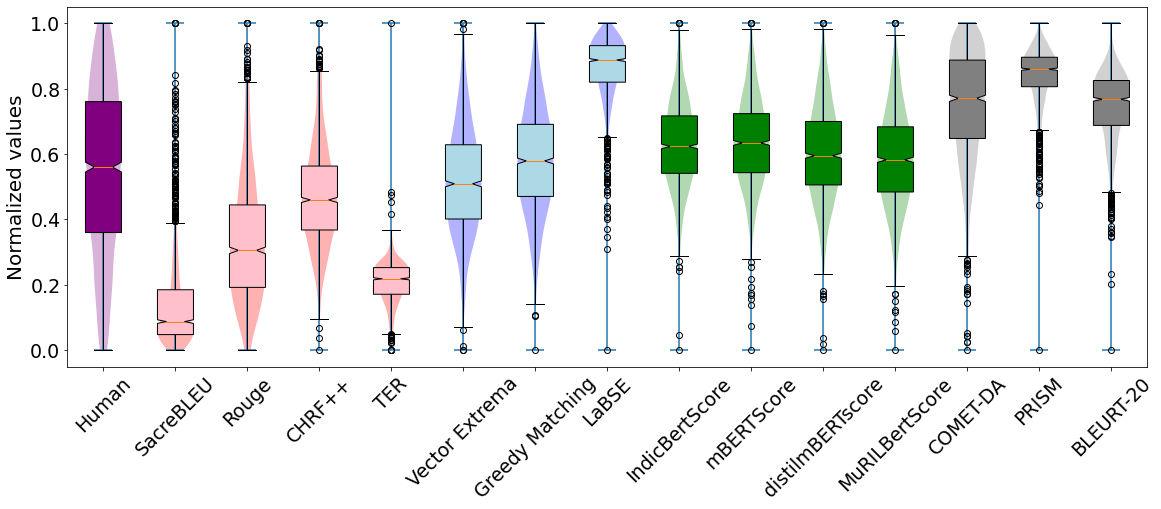}
    \caption{Spread of the metric scores for Tamil. This plot contains a representative subset of metrics, color-coded based on the category of the metric i.e. \textcolor{pinkbox}{pink} for overlap-based metrics, \textcolor{bluebox}{blue} for embedding-based, \textcolor{greenbox}{green} for BERTScore based and \textcolor{greybox}{grey} for trained metrics. We can see that the metric scores are skewed in general while the human scores are not.}
    \label{fig:spread1}
\end{figure}
 \input{tab_fluency_corr.tex}

\subsection{System-level Evaluation}
\label{sec:ranking_metrics}

Table \ref{tab:sys_level_corrs} shows the Pearson and Kendall-tau correlations at the system-level following \citet{10.1162/COLI_a_00123}. Since Kendall-tau is based on pairwise score comparisons, it reflects a common ranking use case and is more reliable for system-level correlations. 
The metric rankings remain consistent across both granularities, with more variability observed in the segment-level task. Similar to the segment-level correlations, trained metrics show the highest correlations across all languages. COMET shows the highest correlations, followed by BLEURT-20. Although on the segment level COMET-QE was not at par with the COMET reference-based metrics, for system ranking the reference-free COMET-QE metrics show high correlations and are well suited for ranking system pairs. Although \citet{kocmi-etal-2021-ship} already observed this for other languages, with the help of our dataset and experiments we are able to provide empirical evidence to confirm this for Indian languages.


\subsection{Spread of Metric Scores}
\label{sec:metrics_spread}
While the correlations of metrics are good, we still find that the range of metric scores is skewed. That is, most of the metrics do not utilise their entire scoring range, and often provide scores in a narrow range. 
This skew in the spread hinders the interpretability of the scores provided by the metric.
For example, SacreBLEU has a scoring range between 0 to 100. However, the scores are almost always in the lower half of the scoring range as seen in Figure \ref{fig:spread1} containing the spread of normalised scores of each metric\footnote{Some of the metrics, such as the trained metrics and edit-distance-based metrics, are not bounded to a scoring range. We normalize such metrics using their maximum and minimum values in the current dataset.}. This is not a case of an issue with the data being always poor as we can see in Figure~\ref{fig:spread1} that the human scores for Tamil show a spread through-out the scale. 
On the other hand, the embedding-based metrics, which use cosine similarity, have a theoretical maximum of 1 and minimum of 0; however, the scores are concentrated at the higher end of the scale, rendering the individual scores uninterpretable despite decent correlations.

\subsection{Correlations Conditioned on Error Type}
\label{sec:corr_flu_ade}
\citet{DBLP:conf/acl/MathurBC20_tangled, sai-etal-2021-perturbation} show that correlations do not convey the true picture and it is important to perform in-depth analysis to understand the true ability of the metrics. Hence we perform the following experiment to examine the performance of metrics on the two primary error categories in the MQM framework, i.e, fluency and accuracy. We select those annotated segments that contain only a single error type in order to clearly separate the two error types. This gives us two MQM data subsets, one containing only fluency errors and the other only accuracy errors. Since the dataset size could be different, we control for the size by sampling an equal number of segments from both sets. Figure~\ref{fig:flu_acc_corr} contains the correlation values for the various metrics. Splitting the dataset based on the error types shows a more nuanced picture. The majority of the metrics show a higher correlation with human scores when only accuracy errors are annotated. This implies that the metrics are able to capture the accuracy errors well but fail on fluency-based errors. We hope that future works on designing better evaluation metrics for Indian languages focus more on developing metrics that can capture fluency-based errors.

\input{indic-comet.tex}







\section{Conclusion}
We present a large-scale MQM dataset consisting of 7000 fine-grained annotations, spanning 5 Indian languages and 7 MT systems, for evaluating machine translation metrics for Indian languages. With the help of this dataset, we show that the current pre-trained metrics outperform the overlap-based metrics (\S\ref{sec:results_1}) in terms of correlations with the human scores. Additionally, we also perform an in-depth study (\S\ref{sec:corr_flu_ade}) to identify the drawbacks of the current metrics. We then use our dataset to train an Indic specific COMET metric that outperforms existing metrics in terms of both correlations and robustness scores (\S\ref{sec:indiccomet_eval}).
We hope that our dataset and analysis will help promote further research in Indic MT evaluation.

\section{Acknowledgements}
We thank the Ministry of Electronics and Information Technology (MeitY), Government of India, for setting up the ambitious Digital India Bhashini Mission with the goal of advancing Indian language technology. The annotators and language experts who worked on this project were supported by the generous grant given by Digital India Bhashini Mission to IIT Madras to serve as the Data Management Unit for the mission. We are indebted to Shri Nandan Nilekani and Shrimati Rohini Nilekani for supporting our work through generous grants from EkStep Foundation and Nilekani Philanthropies. These grants were used for (i) supporting many of the students, research associates and developers who worked on this project, (ii) fulfilling many of our compute needs and (iii) recruiting project managers to oversee the massive pan-India data collection activity undertaken as a part of this work. We thank the Centre for Development and Advancement of Computing, Pune (CDAC Pune) for access to its Param Siddhi super-computer which was used for mining bitext pairs at scale. We thank the Google India Ph.D. Fellowship Program and the Prime Minister’s Fellowship Scheme for Doctoral Research for supporting Ananya Sai.

\section{Limitations}
The approach to collect our dataset is expensive and laborious. This along with the dependence on expert annotators makes the transfer of such an approach challenging for other low-resource languages. We however, find this a necessary endeavor to develop initial resources that can help provide a starting point to extend access to more languages and iteratively improve research, technologies and services across languages.

\section{Ethical Considerations}
For the human annotations on the dataset, the language experts were paid a competitive monthly salary to help with the task. The salary was determined based on the skill set and experience of the expert and adhered to the norms of the government of our country. The dataset has no harmful content. The annotations are collected on a publicly available dataset and will be released publicly for future use. All the datasets created as part of this work will be released under a CC-0 license\footnote{\url{https://creativecommons.org/publicdomain/zero/1.0}} and all the code and models will be release under an MIT license\footnote{\url{https://opensource.org/licenses/MIT}}.


\bibliography{anthology,custom}
\bibliographystyle{acl_natbib}

\appendix
\input{appendix.tex}

\end{document}

%% file: s2RelatedWork.tex
\section{Related Work}
\paragraph{Meta-evaluation studies:}
Evaluation metrics have been under intense scrutiny in order to establish their reliability. \citet{DBLP:conf/acl/MathurBC20_tangled} discuss that studying evaluation metrics needs to be a meticulous task by showing many potential issues and oversights that could lead to wrong conclusions.
Other works focus on extending the resources for meta-evaluations \cite{sai-etal-2021-perturbation, DBLP:journals/corr/abs-2210-13746_perturb} and different genres \cite{DBLP:conf/lrec/WeesBM18_genres}. 
While most of these works focus on English, there are works that evaluate the efficacy of metrics on other languages such as German, Chinese, Spanish, etc. \cite{rivera-trigueros-olvera-lobo-2021-building,freitag2021experts}. On the other hand, we focus on Indian languages, which have not received much attention.

\paragraph{Collecting human annotations:}
Meta-evaluation studies rely heavily on human-annotated translations of various systems. Since humans are better at providing relative ranking (i.e., comparing the qualities of 2 or more items) rather than providing absolute scores to quantify the quality of an item, WMT15-17 collected Relative Rankings \cite{bojar-etal-2015-findings, bojar-etal-2016-findings, bojar-etal-2017-findings}. However, since they require a quadratic number of ratings, Direct Assessment (DA) scores, which are \textit{quality assessment} scores over each output on a scale of 0-100, are easier and faster to collect \cite{kocmi-etal-2021-ship}. More recently, the Multidimensional Quality Metric (MQM) approach for collecting human judgments was adopted for Machine Translation by \citet{freitag2021experts}. They obtained annotations from professional raters with MQM training, which \citet{clark-etal-2021-thats} recommend. On a related note, \citet{DBLP:journals/mt/KlubickaTS18_croatian} conduct human studies for Croatian, whereas \citet{fabbri-etal-2021-summeval} followed systematic approaches to collect and provide multidimensional scores for other tasks such as summarization.

%% file: s3IndicEvalDataset.tex
\section{Indic-MT Eval Dataset}
Following \citet{freitag2021experts}, we collect MQM annotations for 5 Indian languages, i.e., Tamil (ta), Gujarati (gu), Hindi (hi), Marathi (mr), and Malayalam (ml).
We sample 200 sentences from the FLORES-101 dataset \citep{goyal-etal-2022-flores} and obtain the translation outputs from 7 machine translation systems (\S\ref{sec:mt_sys}) for each of the 5 Indian languages.

\begin{figure*}[t]
    \centering
    \includegraphics[scale=0.3]{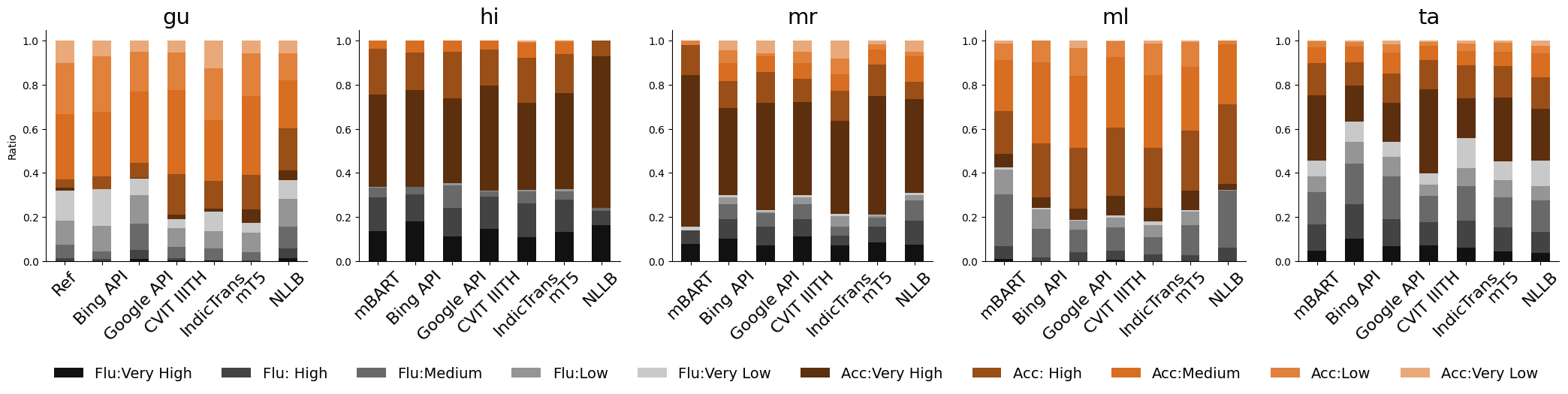}
    \caption{Distribution of the various error types and severity across 5 Indian languages. Darker the shade the more severe the errors. The error types are categorized under two categories Fluency (Flu.) and Accuracy (Acc.). }
    \label{ratioplot}

\end{figure*}

\subsection{MT Systems Considered}
\label{sec:mt_sys}
We use state-of-the-art models to obtain translation outputs in the 5 languages. These include English-XX translation outputs obtained from open-sourced pre-trained models like  mBART \citep{liu-etal-2020-multilingual-denoising}, mT5 \citep{xue-etal-2021-mt5}, IndicTrans \citep{10.1162/tacl_a_00452}, cvit \citep{philip-etal-2019-cvits}, NLLB \citep{costa2022no}, as well as outputs obtained\footnote{Collected in February 2022} from Microsoft Azure Cognitive Services API\footnote{\href{https://learn.microsoft.com/en-us/azure/cognitive-services/translator/reference/v3-0-reference}{Bing API}} and Google translation API\footnote{\href{https://cloud.google.com/translate/docs/reference/rpc/google.cloud.translate.v2}{Google API}} (additional details in Appendix~\ref{app:system_details}). Note that for Gujarati, we find all mBART outputs to be unintelligible and filled with a mixture of characters from several languages. We hence re-allocate the budget corresponding to these sentences for Gujarati to collect annotations on the references instead. Similar to the findings of \citet{clark-etal-2021-thats}, we observe that the references are not always perfect, and these sentences also have errors. However, we find that these errors are often of lower severity.

\begin{figure}[h]
        \centering
        \includegraphics[width= 0.45\textwidth]{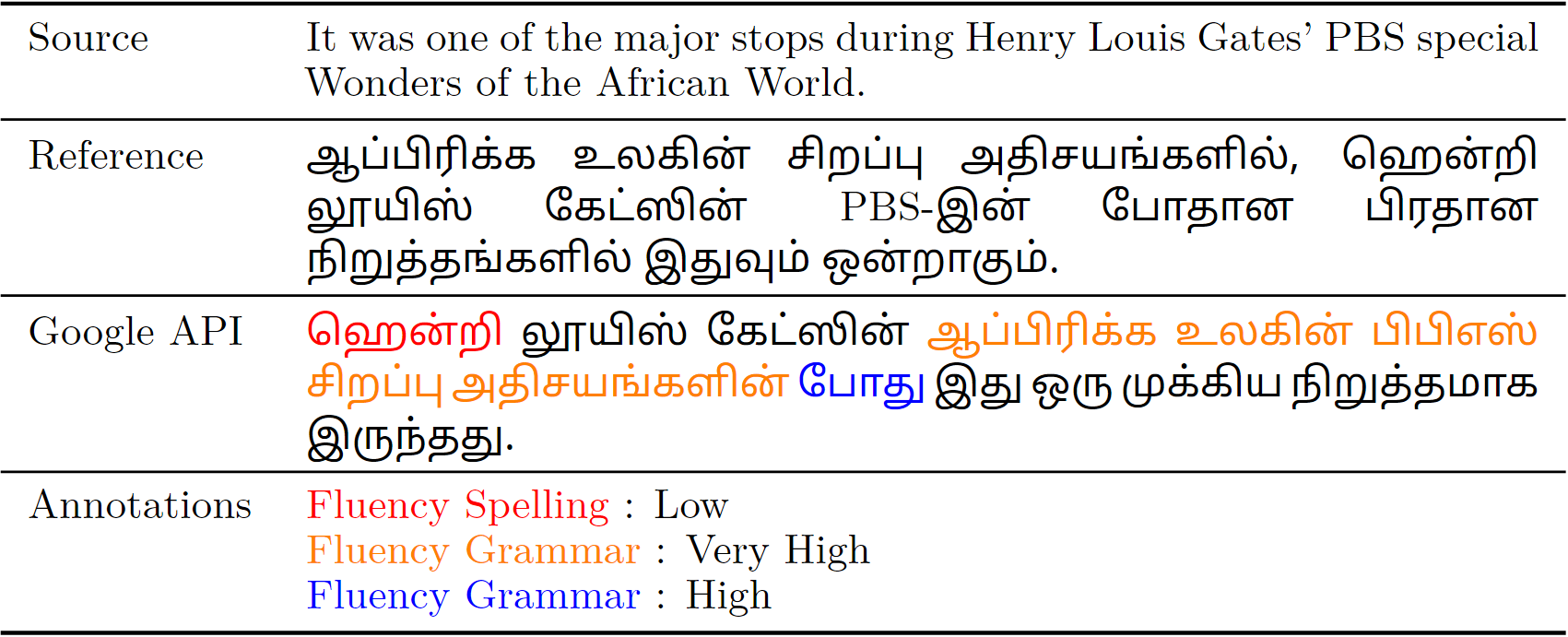}
        \caption{Source, reference and translated output with error spans as demarcated by the annotator.}
        \label{fig:error_categ}
\end{figure}

\subsection{Methodology}
We adopt the MQM-framework \cite{mqm_lommel2014multidimensional} for collecting human annotations on the data at the segment level. In general, a segment may contain one or more sentences. Bilingual language experts, proficient in English and a native language, were employed as annotators for the task of identifying and marking errors in each segment. 
As shown in Figure~\ref{fig:error_categ}, the source segment in English and the translated segment are presented to the annotators, along with provisions to mark up to 5 errors of various categories and severity (\S\ref{sec:analysis}). If there are more than five errors, the annotators are asked to identify only the five most severe ones. In cases where there are more than five severe errors, or if it is not possible to reliably identify distinct errors because the translation is unrelated to the source, then the translation is marked as non-translation, a special category error spanning the entire segment. Depending on the quality of the translation and the errors identified, the annotators were also asked to provide a score out of 25 for each translation after marking all the errors (if any) for that translation. More detailed guidelines are presented in Appendix~\ref{sec:appendix_mqm_guidlines}. 



\subsection{Quality Assurance}
We first performed pilot studies on collecting data via crowd-sourced annotators who are native speakers of the languages we use in this study. In a pilot which directly asked for the final scores, similar to DA scores used in WMT in a few years \cite{DBLP:conf/wmt/BojarGKS16_wmt-16-results, bojar2016ten_years_WMT}, we found the scores to be highly subjective, similar to \citet{clark-etal-2021-thats}. 
We also found that displaying the reference translations, which are not always perfect, was biasing the annotators to ignore some errors.  Another pilot task involved MQM instead of DA scores in the same crowd-sourced setting. However, we found it difficult to achieve consistency in annotations with crowd-sourced raters. We tried the following strategies to improve the quality (i) We provided the same set of segments to 3 annotators per language and then organized a discussion among them to resolve disagreements. The idea was to eventually converge to a consistent marking scheme after a few initial sets of different markings, \cite{DBLP:conf/emnlp/NemaK18_qbleu}.
(ii) We collected annotations from 3 annotators per language and provided all the 3 annotations to a different fourth annotator to aggregate them.
However, neither yielded fruitful results in terms of agreement with MQM annotations.

\begin{figure}[t]
    \centering
    \includegraphics[width=0.9\linewidth]{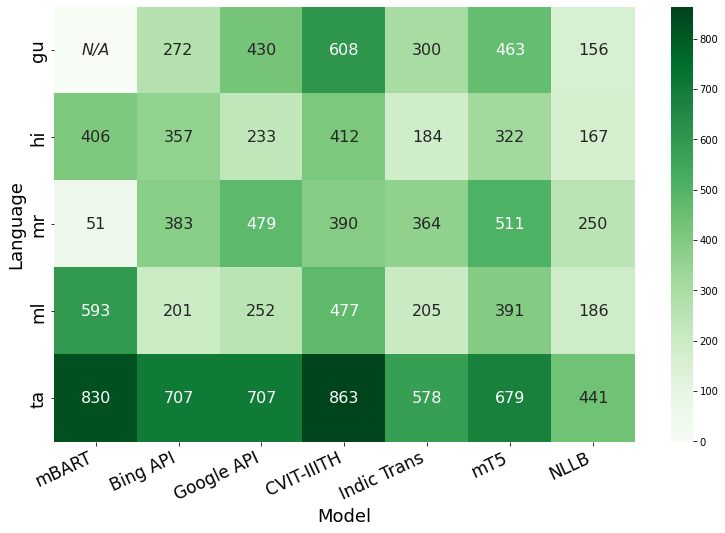}
    \caption{Distribution of the total number of errors per model across each language in consideration.}
    \label{fig:no_errors_per_model}
\end{figure}

Finally, we employed language experts who have experience in translation tasks and observe that we were able to achieve better consistency among annotators. We use the first 50 sentences (sampled randomly from various models and of various lengths) as a pilot to help the annotator get an idea of the variety and kind of translations in the dataset.  Note that MQM-style annotations use a formula to automatically compute scores for each segment based on the errors identified. The score, $s$, for each segment with a set of identified errors, $E$, is given by $s = 25 - \sum_{i \in E} w_i*e_i$, where $w_i$ is the penalty associated with the severity of the error and $e_i$ is the penalty associated with the type of error. Appendix \ref{sec:appendix_mqm_guidlines} provides more details on the penalties used for the different error types and severities, following \citet{mqm_lommel2014multidimensional}. 

In addition to the formula-based score, we also ask the annotator to provide an overall score after marking the errors for that segment. We then verify the correlations between the formula-based scores and the scores provided by the annotator and found them to be highly correlated (i.e., $>0.7$ Kendall-tau correlation) for all languages.
In order to compute the Inter Annotator agreement score, we sample 200 segments for each language and compute the Kendall-tau correlation between the scores given by two annotators. For all the languages, we observe high correlation scores of 0.61, 0.57, 0.55, 0.538, and 0.52 for Malayalam, Gujarati, Tamil, Hindi, and Marathi respectively.




\subsection{Analysis}
\label{sec:analysis}
    \paragraph{Distribution of Error types:} Following the MQM guidelines and prior work on MQM \cite{freitag2021experts}, we have 13 categories of errors, including 4 sub-categories under fluency and 5 under accuracy, style error, source error, non-translation (a special case to mark segments that are extremely poor translations or have more than 5 high severity errors) and an \textit{`other'} category for any error types that are not accounted for in the list of error types. The error types are listed in Appendix~\ref{sec:appendix_mqm_guidlines}. On all languages except Tamil, we found \textit{`Accuracy Mistranslation'} to have the highest error count among all error types. 
    More generally, on average, the machine translation models today primarily err on accuracy-based errors and make fewer fluency-based mistakes as seen in Figure~\ref{ratioplot}. 
    
    \paragraph{Severity of Errors:} We plot all the fluency and accuracy errors graded by error severities for all languages in Figure~\ref{ratioplot}. As depicted, there are 5 error severity types: \textit{Very High, High, Medium, Low}, and \textit{Very Low}.  For all the Indo-Aryan languages (gu, hi and mr), the majority of the errors observed are accuracy-based errors. For Tamil, a Dravidian language, we find the accuracy errors and fluency errors in almost equal proportions. We find Malayalam, another Dravidian language, to have more accuracy errors than fluency errors, with majority medium-severity errors, as shown in Figure 1.

\paragraph{MT systems:} Figure~\ref{fig:no_errors_per_model} shows the total number of errors per model (inclusive of all severities) for each language. We find that the recent MT models (NLLB, IndicTrans) have fewer errors compared to the relatively older models (CVIT). 
Table~\ref{tab:avgcompscore} in Appendix provides a more detailed picture which also inherently takes into account the severities of the errors. It shows the average score of each system computed as the mean of the human scores obtained on all the outputs from that system. We find that IndicTrans model, which focuses on Indian languages, has the highest scores on Hindi, Malayalam and Tamil. NLLB is the best performing model on Marathi and Bing API for Gujarati. Considering the average performance across all languages, the best performing models in descending order are IndicTrans, NLLB, Google API, Bing API, mT5, CVIT, and mBART.

%% file: s4ExperimentalSetup.tex
\section{Experimental Setup}
In this section, we discuss the various evaluation metrics under consideration (\S\ref{sec:metrics}) and evaluating strategies (\S\ref{sec:strat}) followed.

\input{tab_corrs_mqm_metrics}

\subsection{Evaluation Metrics Used for MT}
\label{sec:metrics}
We consider the most popular metrics being used in \citet{wmt-2021-machine, barrault-etal-2020-findings} along with other variants to suit the languages under consideration. In total, we study 16 metrics belonging to different classes \citep{sai2022survey} of either word overlap-based, embedding-based, or trained metrics. 
\begin{itemize}
    \item In the word overlap-based category, we consider 
(i) BLEU \cite{bleu}, (ii) SacreBLEU \cite{sacrebleu}, (iii) ROUGE \cite{rouge}, (iv) chrF++ \cite{DBLP:conf/wmt/Popovic17_chrFpp}, (v) TER \cite{Snover06astudy_ter}. 
\item For the embedding-based metrics, we use (i) Vector Extrema (VE) \cite{vectorextrema}, (ii) Greedy Matching (GM) \cite{greedymatch}, (iii) Embedding Averaging (EA) \cite{landauer1997solution_vector_averaging1}, (iv) LabSE \cite{labse} \& (v) LASER \cite{laser} embeddings and (vi) BERTScore \cite{DBLP:conf/iclr/ZhangKWWA20_BERTscore}. 
\item For computing BERTScore, in addition to the official implementation, which uses mBERT, we also consider other variants that use BERT models trained on Indian languages, namely IndicBERT \cite{indicbert} and MuRIL \cite{muril}.
\item The end-to-end trained metrics we consider are (i) PRISM \cite{prism}, (ii) BLEURT \cite{DBLP:conf/acl/SellamDP20_bleurt} and (iii) COMET variants \citep{rei-etal-2020-comet}.
%
\end{itemize}

\subsection{Meta Evaluation}
\label{sec:strat}
For evaluating the evaluation metrics we measure how well the metrics correlate with human judgments on two granularities i.e.: segment-level and system-level. We use Pearson correlation ($\rho$) which measures the linear correlation between two sets of data and Kendall's Tau ($\tau$) to measure the ordinal association between two quantities. 

%% file: tab_corrs_mqm_metrics.tex
\begin{table*}[t!]
\footnotesize
\small 
\centering
\resizebox{\textwidth}{!}{
\begin{tabular}{l|cc|cc|cc|cc|cc|cc}
\toprule
            & \multicolumn{2}{c|}{\bf gu} & \multicolumn{2}{c|}{\bf hi} & \multicolumn{2}{c|}{\bf mr} & \multicolumn{2}{c|}{\bf ml} & \multicolumn{2}{c|}{\bf ta} & \multicolumn{2}{c}{\bf Average} \\
\multirow{-2}{*}{\bf Metric}      & $\rho$          & $\tau$           & $\rho$          & $\tau$           & $\rho$          & 	$\tau$           & $\rho$          & $\tau$           & $\rho$          & $\tau$          & $\rho$           & $\tau$           \\ 
\midrule
BLEU 1      & 0.364      & 0.255      & 0.266      & 0.187      & 0.228      & 0.148      & 0.393      & 0.331      & 0.316      & 0.213      & 0.314       & 0.227      \\
BLEU 2      & 0.329      & 0.247      & 0.280      & 0.192      & 0.190      & 0.135      & 0.331      & 0.302      & 0.291      & 0.205      & 0.284       & 0.216      \\
BLEU 3      & 0.294      & 0.234      & 0.265      & 0.186      & 0.134      & 0.119      & 0.250      & 0.271      & 0.227      & 0.182      & 0.234       & 0.198      \\
BLEU 4      & 0.235      & 0.215      & 0.245      & 0.171      & 0.091      & 0.103      & 0.180      & 0.246      & 0.171      & 0.168      & 0.184       & 0.181      \\
SacreBLEU   & 0.293      & 0.239      & 0.255      & 0.168      & 0.164      & 0.132      & 0.274      & 0.298      & 0.244      & 0.189      & 0.246       & 0.205      \\
ROUGE-L     & 0.350      & 0.251      & 0.295      & 0.204      & 0.206      & 0.132      & 0.376      & 0.322      & 0.308      & 0.206      & 0.307       & 0.223      \\
chrF++      & \bf 0.408      & \bf 0.287      & \bf 0.299      & \bf 0.205      & 0\bf .260      & \bf 0.170      & \bf 0.411      & \bf 0.338      & \bf 0.361      & \bf 0.250      & \bf 0.348       & \bf 0.250      \\
TER         & 0.304      & 0.237      & 0.263      & 0.196      & 0.203      & 0.135      & 0.343      & 0.307      & 0.272      & 0.199      & 0.277       & 0.215      \\
\midrule
EA          & 0.331      & 0.181      & 0.086      & 0.066      & 0.143      & 0.054      & 0.397      & 0.301      & 0.203      & 0.149      & 0.232       & 0.150      \\
VE          & 0.380      & 0.265      & \bf 0.274      & \bf 0.183      & 0.234      & 0.153      & 0.412      & 0.331      & 0.337      & 0.227      & 0.327       & 0.232      \\
GM          & 0.394      & 0.266      & 0.234      & 0.162      & 0.241      & 0.147      & \bf 0.426      & \bf 0.338      & \bf 0.382      & 0.264      & 0.335       & 0.235      \\
LASER embs  & 0.094      & 0.156      & 0.135      & 0.123      & 0.159      & 0.069      & 0.357      & 0.295      & 0.126      & 0.099      & 0.174       & 0.148      \\
LabSE embs  & \bf 0.504      & \bf 0.319      & 0.149      & 0.185      & \bf 0.319      & \bf 0.204      & 0.416      & 0.337      & 0.339      & \bf 0.286      & \bf 0.345       & \bf 0.266      \\
\midrule
mBERT       & 0.448      & 0.297      & 0.337      & 0.231      & \bf 0.301      & \bf 0.194      & 0.462      & 0.367      & 0.413      & 0.281      & 0.392       & 0.274      \\
distilmBERT & 0.431      & 0.289      & 0.316      & 0.220      & 0.281      & 0.181      & \bf 0.465      & \bf 0.371      & \bf 0.415      & 0.278      & 0.382       & 0.268      \\
IndicBERT   & 0.456      & 0.308      & 0.346      & 0.235      & 0.281      & 0.182      & 0.440      & 0.357      & 0.402      & 0.282      & 0.385       & 0.273      \\
MuRIL       & \bf 0.465      & \bf 0.322      & \bf 0.353      & \bf 0.243      & 0.292      & 0.184      & 0.449      & 0.369      & 0.410      & \bf 0.290      & \bf 0.394       & \bf 0.282      \\
\midrule
PRISM       & 0.114      & 0.024      & 0.178      & 0.124      & 0.131      & 0.084      & 0.089      & 0.064      & -0.040     & -0.040     & 0.094       & 0.051      \\
BLEURT-20   & 0.509      & 0.371      & 0.296      & 0.300      & 0.409      & 0.286      & 0.496      & 0.390      & 0.491      & 0.374      & 0.440       & 0.344      \\

COMET-QE-DA 
& 0.417 & 0.324 
& 0.535 & \bf 0.404
& 0.551 & 0.430
& 0.386 & 0.341
&0.531 & 0.391 
& 0.414 & 0.378\\

COMET-QE-MQM 
& 0.387 & 0.309
& 0.590 & 0.403
& 0.577 & 0.392 
& 0.438 & 0.392
& 0.571 & 0.399
& 0.513 & 0.379
\\

COMET-DA    & \bf 0.557     & \bf 0.403     & \bf 0.581     & 0.390     &  0.426    &  0.306      & \textbf{0.531}      & \textbf{0.419}      &  0.529      &  0.412      & 0.525      & 0.386      \\

COMET-MQM   & 0.465    & 0.360      & 0.529      &   0.370    &  \bf 0.686     &  \bf 0.459   & 0.508       &   0.392    & \bf 0.597     & \bf  0.432    & \bf 0.557      & \bf 0.402     \\

\bottomrule
\end{tabular}
}
\caption{Segment-level Pearson ($\rho$) and Kendall tau ($\tau$) correlations of different metrics. The best metric correlation amongst each metric category \cite{sai2022survey} in \textbf{bold}. We observe that COMET-MQM is the best-performing metric overall for all languages in consideration. All correlations are significant ($p< 0.05$). }
\label{tab:corr_formula_metrics}
\end{table*}

%% file: tab_sys_level_corrs.tex
\begin{table*}[t!]
\footnotesize
\small 
\centering
\resizebox{\textwidth}{!}{
\begin{tabular}{l|cc|cc|cc|cc|cc}
\toprule
            & \multicolumn{2}{c|}{\bf gu} & \multicolumn{2}{c|}{\bf hi} & \multicolumn{2}{c|}{\bf mr} & \multicolumn{2}{c|}{\bf ml} & \multicolumn{2}{c}{\bf ta}  \\
\multirow{-2}{*}{\bf Metric}      & $\rho$          & $\tau$           & $\rho$          & $\tau$           & $\rho$          & 	$\tau$           & $\rho$          & $\tau$           & $\rho$          & $\tau$              \\ 
\midrule
BLEU 1      
& 0.927$^{*}$  & 0.600  
& 0.684  & 0.429  
& \bf 0.949$^{*}$  & 0.143  
& \bf 0.913$^{*}$  & 0.619  
& 0.698  & 0.429  \\

BLEU 2      
& 0.922$^{*}$  & 0.600  
& 0.697  & 0.524  
& 0.922$^{*}$  & 0.143  
& 0.885$^{*}$  & 0.619  
& 0.714  & 0.619  \\

BLEU 3      
& \bf 0.930$^{*}$  & 0.600  
& 0.687  & 0.524  
& 0.891$^{*}$  & 0.143  
& 0.829$^{*}$  & 0.619  
& 0.674  & 0.619  \\

BLEU 4      
& 0.914$^{*}$  & 0.600  
& 0.651  & 0.429  
& 0.793$^{*}$  & 0.143  
& 0.772$^{*}$  & 0.619 
& 0.598  & 0.524  \\

SacreBLEU   
& 0.926$^{*}$  & 0.600  
& 0.648  & 0.429  
& 0.912$^{*}$  & 0.143  
& 0.849$^{*}$  & 0.619 
& 0.656  & 0.619  \\

ROUGE-L     
& 0.928$^{*}$  & 0.600  
& 0.741  & 0.524  
& \bf 0.949$^{*}$  & 0.143  
& 0.909$^{*}$  & 0.619 
& 0.697  & 0.524  \\

chrF++      
& 0.923$^{*}$  & 0.600  
& 0.67   & 0.429  
& 0.9$^{*}$    & 0.429  
& 0.895$^{*}$  & 0.524  
& \bf 0.756$^{*}$  & 0.619  \\

TER         
& -0.931$^{*}$ & -0.600 
& -0.757$^{*}$ & -0.524 
& -0.977$^{*}$ & -0.143 
& -0.911$^{*}$ & -0.619 
& -0.696 & -0.619 \\

\midrule
EA          
& 0.927$^{*}$  & 0.600  
& 0.547  & 0.411  
& 0.968$^{*}$  & 0.238  
& 0.919$^{*}$  & 0.586 
& 0.739  & 0.429  \\

VE          
& \bf 0.952$^{*}$  & 0.733 
& 0.654  & 0.524  
& 0.967$^{*}$  & 0.143  
& \bf 0.958$^{*}$  & 0.619 
& 0.766$^{*}$  & 0.524  \\

GM          
& 0.942$^{*}$  & 0.733  
& 0.636  & 0.524  
& \bf 0.977$^{*}$  & 0.143  
& 0.949$^{*}$  & 0.619 
& \bf 0.777$^{*}$  & 0.524  \\

LASER       
& 0.273  & 0.067  
& 0.372  & 0.143  
& 0.797$^{*}$  & 0.048  
& 0.873$^{*}$  & 0.429  
& 0.67   & 0.333  \\

LabSE       
& 0.931$^{*}$  & 0.600  
& 0.253  & 0.048  
& 0.968$^{*}$  & 0.238  
& 0.823$^{*}$  & 0.333 
& 0.725  & 0.429  \\
\midrule

mBERT       
& 0.947$^{*}$  & 0.600  
& 0.705  & 0.524  
& \bf 0.978$^{*}$  & 0.143  
& 0.940$^{*}$   & 0.683  
& 0.798$^{*}$  & 0.524  \\

distilmBERT 
& 0.945$^{*}$  & 0.600  
& 0.629  & 0.429  
& 0.976$^{*}$  & 0.143  
& \bf 0.946$^{*}$  & 0.683$^{*}$  
& \bf 0.825$^{*}$  & 0.524  \\

IndicBERT   
& 0.949$^{*}$  & 0.733 
& 0.747  & 0.524  
& 0.971$^{*}$  & 0.143  
& 0.938$^{*}$  & 0.619 
& 0.758$^{*}$  & 0.524  \\

MuRIL       
& \bf 0.957$^{*}$  & 0.733
& 0.742  & 0.524  
& 0.976$^{*}$  & 0.143  
& 0.926$^{*}$  & 0.619
& 0.777$^{*}$  & 0.524  \\

\midrule
PRISM       
& 0.810  & 0.467  
& 0.583  & 0.238  
& 0.979$^{*}$  & 0.238  
& 0.877$^{*}$  & 0.619 
& 0.611  & 0.238  \\

BLEURT-20   
& 0.978$^{*}$  & \bf 1.000$^{*}$
& 0.582  & 0.714$^{*}$ 
& \bf 0.993$^{*}$  & 0.619 
& 0.952$^{*}$  & 0.39   
& \bf 0.927$^{*}$  & \bf 0.905$^{*}$  \\

COMET-QE-DA
&0.852$^{*}$ & 0.866$^{*}$
& 0.878$^{*}$ & 0.714$^{*}$
& 0.854$^{*}$ & 0.714$^{*}$
& \bf 0.986$^{*}$ & \bf 0.809$^{*}$
& 0.911$^{*}$ & 0.714$^{*}$ \\

COMET-QE-MQM
& 0.657 & 0.733
& 0.831$^{*}$ & 0.809$^{*}$
& 0.971$^{*}$ &0.619
& 0.798$^{*}$ &0.428
& 0.892$^{*}$ & 0.714$^{*}$\\

COMET-DA    
& \bf 0.986$^{*}$  & \bf1.000$^{*}$  
& \bf 0.970$^{*}$   & \bf 1.000$^{*}$    
& 0.994$^{*}$  & 0.781$^{*}$ 
& 0.936$^{*}$  & 0.333  
& 0.868$^{*}$  & 0.619$^{*}$
\\

COMET-MQM
& 0.932$^{*}$ & 0.733
& 0.759$^{*}$ & 0.809$^{*}$
& 0.991$^{*}$ & \bf 0.904$^{*}$
& 0.953$^{*}$ & 0.523
& 0.892 & 0.714\\

\bottomrule
\end{tabular}
}
\caption{System-level Pearson ($\rho$) and Kendall-tau ($\tau$) correlations of different metrics. The best performing metric in each category in \textbf{bold}. ($^{*}$) signifies that the correlation value is significant ($p< 0.05$).}
\label{tab:sys_level_corrs}
\end{table*}

%% file: tab_fluency_corr.tex
\begin{figure*}[t]
    \centering
    \includegraphics[width=\textwidth]{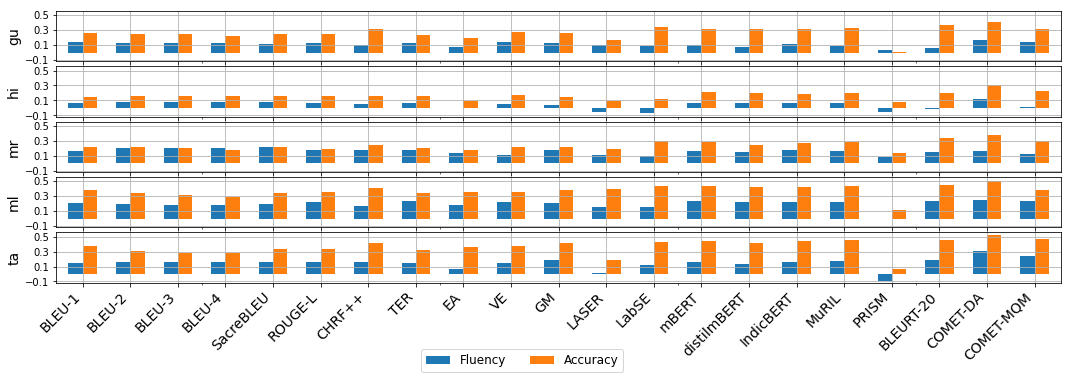}
    \caption{Kendall-tau ($\tau$) correlations of the different metrics with the two MQM subsets (Fluency and Accuracy) across the 5 languages. We can observe that all the metrics on average correlate better with the human scores when only accuracy errors are annotated compared to fluency errors }
    \label{fig:flu_acc_corr}
\end{figure*}

%% file: indic-comet.tex
\begin{table*}[t]
\small
\centering
\resizebox{\textwidth}{!}{
\begin{tabular}{lcccccccccccc}
\toprule
\multirow{2}{*}{\bf Metrics}         
& \multicolumn{2}{c}{\bf gu} & 
\multicolumn{2}{c}{\bf hi} & 
\multicolumn{2}{c}{\bf mr}  & 
\multicolumn{2}{c}{\bf ml} & 
\multicolumn{2}{c}{\bf ta} &
\multicolumn{2}{c}{\bf Avg.} 
\\
& $\rho$    & $\tau$   & $\rho$    & $\tau$   & $\rho$    & $\tau$ & $\rho$    & $\tau$   & $\rho$    & $\tau$   & $\rho$    & $\tau$ 
\\
\midrule
COMET-DA
& \bf 0.487 
& 0.359 
& 0.380 
& 0.319 
& 0.422 
& 0.302 
& 0.529 
& \bf 0.421 
& 0.525 
& 0.410 
& 0.469
& 0.362
\\
COMET-MQM
& 0.422 
& 0.346 
& 0.528 
& 0.370 
& 0.455 
& 0.314 
& 0.493 
& 0.380 
& 0.588 
& 0.429 
& 0.497
& 0.367\\
\midrule
IndicCOMET$_{XLM}$
& 0.437 
& 0.353
& 0.609
& 0.397
& 0.413
& 0.311
& 0.559
& 0.418
& 0.585
& 0.426
& 0.521
& 0.381 \\
IndicCOMET$_{DA}$
& 0.431
& 0.339
& 0.554
& 0.384
& 0.436
& 0.310
& 0.526
& 0.410
& 0.587
& 0.433
& 0.507
& 0.375 
\\
IndicCOMET$_{MQM}$
& 0.446
& \bf 0.360
& \bf 0.616
& \bf 0.419
& \bf 0.463
& \bf 0.331
& \bf 0.566
& 0.416
& \bf 0.597
& \bf 0.441
& \bf 0.537
& \bf 0.393
\\
\bottomrule
\end{tabular}
}
\caption{Correlations values of Indic-COMET. The highest value in each column in \textbf{bold} ( $p < 0.05$). \textit{XLM, DA} and \textit{MQM} imply that the IndicCOMET weights were initialized from the XLM-R, COMET-DA, and COMET-MQM checkpoints respectively. Initializing the metric with the COMET-MQM shows the highest correlations on average. }
\label{tab:trained_comet_scores}
\end{table*}

\section{Indic COMET}
\label{sec:indic_comet}
Having analyzed various metrics, we fine-tune the best performing metric -- COMET -- using our MQM dataset (\S\ref{sec:comet_training}) and show that the new fine-tuned metric not only outperforms the COMET metric on the majority of the languages but also is more robust to perturbations (\S\ref{sec:indiccomet_eval}). Additionally, we also test the zero-shot evaluation ability of the Indic-COMET metric in \S\ref{sec:indiczeroshot}.

\subsection{Training}
\label{sec:comet_training}
We build our metric with the architecture of COMET \cite{rei-etal-2020-comet}. We use the Estimator model, which uses XLM-RoBERTa \cite{conneau2019unsupervised} backbone to encode the source, hypothesis, and reference. We use the same training process and hyper-parameters as COMET for a fair comparison (additional details in Appendix~\ref{app:indic_comet}). Following \citet{rei-etal-2021-references}, we experiment with initializing the model with different checkpoints, namely, XLM-R, COMET-DA, and COMET-MQM, and fine-tune it on our MQM dataset.

\subsection{Evaluation}
\label{sec:indiccomet_eval}
Table~\ref{tab:trained_comet_scores} compares the correlation values of our fine-tuned Indic-COMET with the best-performing COMET baselines. Since no other evaluation datasets for Indian languages are available, we use our own MQM dataset for both training and testing. Hence to perform a throughout evaluation we perform a 3-fold cross-evaluation by splitting the dataset into 3 independent training and testing datasets and report the mean correlation values across the 5 languages in consideration. We observe that Indic-COMET fine-tuned from the COMET-MQM checkpoint shows higher correlations across all languages, compared to the other variants on average. Indic-COMET-MQM outperforms both the COMET baselines on 3 out of the 5 languages and shows higher correlations than COMET-MQM across all languages. The most notable gains are in Hindi. Inspired by recent works on meta-evaluation \cite{kocmi-EtAl:2022:WMT, sai-etal-2021-perturbation}, we also analyze the robustness of metrics on challenge sets. We make use of the challenge set created by \citet{amrhein2022aces} since it contains Indian languages. We use the subset of the dataset that only contains Indian languages and follow \citet{amrhein2022aces} to report performance with Kendall’s tau-like correlations. Indic-COMET-MQM has a correlation score of 0.306 and is more robust than the COMET counterpart which has a score of 0.272. Overall, we observe that fine-tuning the COMET metric on our MQM dataset not only improves correlations with human scores but also increases the robustness to perturbations.


\begin{table}[t]
    \centering
    \small 
    \resizebox{\columnwidth}{!}{
    \begin{tabular}{lccccc}
             \toprule
         \bf Metrics & \bf gu & \bf hi & \bf mr & \bf ml & \bf ta \\
         \toprule
        COMET$_{DA}$            
        & \bf 0.359         
        & 0.319           
        & 0.302        
        & \bf 0.421          
        & 0.410    \\
        COMET$_{MQM}$
        & 0.346
        & 0.370
        & 0.314
        & 0.380
        & 0.429 \\
        \midrule
        IndicCOMET$_{MQM}$
        & 0.355
        & \bf 0.395
        & \bf 0.322
        & 0.394
        & \bf 0.430 \\
        
\bottomrule
    \end{tabular}
    }
    \caption{Kendall-tau ($\tau$) correlations for the zero-shot performance of Indic-COMET$_{MQM}$. Each column corresponds to the language it was not trained on.}
    \label{tab:zs_table}
\end{table}

\subsection{Zero-shot Evaluation}
\label{sec:indiczeroshot}

Since we evaluate only 5 Indian languages, out of the 22 official languages (and over a hundred major languages that are spoken in the country\footnote{https://www.britannica.com/topic/Indian-languages}), we investigate whether the metric has the potential to perform better in other Indian languages as well. In order to test this ability, we finetune on only 4 languages and test on the unseen one. We use the same evaluation setup as discussed in \S\ref{sec:indiccomet_eval}. Table~\ref{tab:zs_table} contains the comparison between the best performing Indic-COMET variant i.e.: Indic-COMET$_{MQM}$ and COMET baselines. We observe that Indic-COMET still outperforms both the COMET baselines on the majority of languages even though it is not trained on the specific Indian languages. It also shows higher correlations than COMET-MQM across all languages. This suggests that collecting annotations for some Indian languages is key for progress in Indic evaluation as it can benefit other low-resource languages too.

%% file: appendix.tex
\section{MQM Guidelines to Annotators, Error types \& Severities}
\label{sec:appendix_mqm_guidlines}

\begin{table*}[t!]
\small
\centering
\resizebox{\textwidth}{!}{
\begin{tabular}{ll|p{0.6\textwidth}}
\toprule
\multicolumn{2}{l|}{\textbf{Error Category}}                          & \textbf{Explanation}                                                                       \\ \midrule
\multicolumn{1}{l}{Accuracy}    & Addition                  & Translation includes information not present in the source.                       \\ 
\multicolumn{1}{l}{}            & Omission                  & Translation is missing content from the source.                                   \\ 
\multicolumn{1}{l}{}            & Mistranslation            & Translation does not accurately represent the source.                             \\ 
\multicolumn{1}{l}{}            & Untranslated text         & Source text has been left untranslated                                            \\ \midrule
\multicolumn{1}{l}{Fluency}     & Spelling                  & Incorrect spelling or capitalization.                                             \\ 
\multicolumn{1}{l}{}            & Grammar                   & Problems with grammar, other than orthography.                                    \\ 
\multicolumn{1}{l}{}            & Register                  & Wrong grammatical register (eg, inappropriately informal pronouns).               \\ 
\multicolumn{1}{l}{}            & Character Encoding        & Characters are garbled due to incorrect encoding. Example: Sink -\textgreater \$ink \\   \midrule
\multicolumn{1}{l}{Terminology Inappropriate} & & Terminology is non-standard or does not fit context.                              \\  \midrule
\multicolumn{1}{l}{Style Awkward} &
   &
  The style of the text does not feel very apt. (Example: 1. The source sentence feels formal like in a newspaper, but the translation doesn’t. 2. Sentences are correct, but simply too long, etc..) \\ \midrule
\multicolumn{1}{l}{Transliteration} &
   &
  If it transliterates instead of translating words/ phrases, where it should not. 
  \\ \midrule

\multicolumn{2}{l|}{Other}                                   & Any other issues.                                                                 \\  \midrule
\multicolumn{2}{l|}{Source Error}                            & An error in the source.                                                           \\  \midrule
\multicolumn{2}{l|}{Non Translation}                         & Impossible to reliably characterize the 5 most severe errors.                     \\ 
\bottomrule
\end{tabular}
}
\caption{Error Hierarchy with corresponding explanations provided to the annotators }
\label{tab:error_hierarchy}

\end{table*}

\begin{table*}[t]
  \small
  \centering
  \begin{tabular}{l|p{0.6\textwidth}}
   \toprule
\textbf{Error Severity} & \textbf{Description}\\
  \midrule
   \multirow{2}{*}{Very High} & Errors that may confuse or mislead the reader due to significant changes in meaning or because they appear in a visible or important part of the content.\\
   \midrule 
   
   \multirow{3}{*}{Very Low} & Errors that don’t lead to loss of meaning and wouldn’t confuse or mislead the reader but would be noticed, would decrease stylistic quality, fluency, or clarity, or would make the content less appealing.\\
   \bottomrule
  \end{tabular}
  \caption{Error Severity End-points Description}
  \label{tab:error_severity}
\end{table*}

The annotators assess translations at the segment level, where a segment may contain one or more sentences. Each translated segment is aligned with a corresponding source segment, and both the source and translated segments are displayed. Table~\ref{tab:error_hierarchy} shows the error hierarchies for all the error types. Each category has severity levels ranging from very high to very low on a 5-point scale of (Very low, Low, Medium, High, and Very-high). Table~\ref{tab:error_severity} shows the descriptions of the end-points of the scale, as shown to the annotators. For computing scores for each segment based on the annotations, we use the following weights/penalties: very low: 1, low: 2, medium: 3, high: 4, very high: 5. Each of the sub-categories under Accuracy, Fluency, Terminology Inappropriate, Style have equal weightage since each of these are accompanied with a corresponding severity marking. Non-translation errors, by definition, elicit a score of 0. Sentences that are marked with a source error are discarded. 

The following guidelines were provided to the annotators: 

\begin{itemize}
    \item 
 Identify all errors within each translated segment, up to a maximum of five. If there are more than five errors, identify only the five most severe. 
    \item 
To identify an error, highlight the relevant span of text using text colors, and select a category/sub-category and severity level from the available options. (The span of text may be in the source segment if the error is a source error or an omission.) 
    \item 
When identifying errors, be as fine-grained as possible. For example, if a sentence contains two words that are each mistranslated, two separate mistranslation errors should be recorded.
    \item 
If a single stretch of text contains multiple errors, (that is, if there are overlapping errors) one only needs to indicate the one that is most severe. If all have the same severity, choose the first matching category listed in the error typology (eg, Accuracy, then Fluency, then Terminology, etc). 
    \item 
There are two special error categories: Source error and Non-translation. Source errors should be annotated separately, highlighting the relevant span in the source segment. A sentence that has a source error need not be scored but  the error in the source segment is to be highlighted.
    \item 
If it is not possible to reliably identify distinct errors because the translation is too badly garbled or is unrelated to the source, then mark a single Non-translation error that spans the entire segment. There can be at most one Non-translation error per segment, which should span the entire segment. No other errors should be identified if Non-Translation is selected.
    \item 
Depending on the quality of the translation and the errors identified, provide a score out of 25 for each translation. Indicate the score in the final score column, after marking all the errors (if any) for that translation.
\end{itemize}

\section{Additional details}

\subsection{MT systems Considered}
\label{app:system_details}
For the mBART we use the Huggingface Transformers \cite{wolf-etal-2020-transformers} for generating the outputs for the various languages. Specifically, we use the \texttt{facebook/mbart-large-50-many\\
-to-many-mmt} model. For mT5 we finetune the pre-trained mT$_{BASE}$ model for the translation task using all existing sources of parallel data provided by \citet{10.1162/tacl_a_00452}. We finetune one model for every language pair. For IndicTrans and CVIT, we use the models released by \citet{10.1162/tacl_a_00452} and \citet{philip-etal-2019-cvits} respectively.

\subsection{Indic COMET Training}
\label{app:indic_comet}
All experiments were conducted using a private infrastructure, which has a carbon efficiency of 0.432 kgCO$_2$eq/kWh. A cumulative of 10 hours of computation was performed on a single RTX A4000 GPU. Total emissions are estimated to be 0.6 kgCO$_2$eq of which 0 percent were directly offset. Estimations were conducted using the \href{https://mlco2.github.io/impact#compute}{MachineLearning Impact calculator} presented in \citet{lacoste2019quantifying}. \\
\begin{table}[h]
    \centering
    \begin{tabular}{ll}
    \toprule
    \bf Hyperparameters & \bf Value \\
    \midrule
    batch size & 16 \\
    dropout & 0.1 \\
    encoder learning rate & 1.0e-05 \\
    encoder model & XLM-RoBERTa \\
    hidden sizes & 3072, 1536 \\
    layer & mix \\
    layerwise decay & 0.95 \\
    learning rate & 3.0e-05 \\
    no. of frozen epochs & 1 \\
    optimizer & AdamW \\
    pool & avg \\

    \bottomrule
    \end{tabular}
    \caption{Hyper-parameters for training the various Indic-COMET model. The initialised model weights are the only difference between all variants; all variants share the same set of hyper-parameters.}
    \label{tab:app_hyper}
\end{table}

\input{violinandscatterplots}

\begin{table}[t]
    \centering
    \small
    \begin{tabular}{l|c|c}
    \toprule
     \textbf{Metric}  & \textbf{Min} & \textbf{Max} \\
     \midrule
     \midrule
     Human  &  0 & 25 \\
     COMET & -1.6 & 1.3 \\
IndicBERT & 0.46 & 1.0 \\
Vector Extrema & 0.3 & 1.0 \\
GM & 0.4 & 1.0 \\
mBERT & 0.56 &  1.0 \\
MurIL & 0.29 & 1.0 \\
TER & 0.0  &  361.1 \\
chrF++ & 1.7 & 100.0 \\
sacreBLEU & 0.0 & 100.0 \\
ROUGE & 0.0 & 100.0 \\
BLEU1 & 0.0 & 100.0 \\
\bottomrule
    \end{tabular}
    \caption{Maximum and minimum values of metrics }
    \label{tab:max_min}
\end{table}

\begin{table*}[t]
\small
\centering
\begin{tabular}{l|c|c|c|c|c|c|c}
\toprule
\multicolumn{8}{c}{\textbf{Average computed human scores for each system}}                                                                                                                                                                                                   \\ \midrule
lang & 
IndicTrans & 
Bing API & 
CVIT-IIITH   & 
Google API & 
mBART & 
mT5  & 
NLLB \\ \midrule

gu & 22.639 & 23.179 & 19.034 & 21.686 & 0.000 & 20.067 & 22.490 \\ 
hi         & 20.120                & 14.405    & 14.962 & 19.484      & 15.703 & 18.012 & 18.445                    \\ 
mr         & 18.484                 & 17.934    & 17.586 & 15.750      & 5.773  & 14.441 & 18.618                    \\ 
ml         & 22.676                 & 22.617    & 17.844 & 21.955      & 17.355 & 20.169 & 21.515                    \\ 
ta         & 17.978                 & 16.516    & 11.933 & 16.651      & 13.522 & 15.994 & 17.578                    \\ \midrule
avg & 20.379                 & 18.930     & 16.272  & 19.105       & 10.471 & 17.737  & 19.729                    \\
\bottomrule
\end{tabular}

\caption{Average human score per system}
\label{tab:avgcompscore}
\end{table*}

For training, we follow the same process as \citet{rei-etal-2020-comet}. We load the pretrained encoder and initialize it with either XLM-Roberta, COMET-DA or COME-MQM weights. During training, we divide the model parameters into two groups: the encoder parameters, that include the encoder model and the regressor parameters, that include the parameters from the top feed-forward network. We apply gradual unfreezing and discriminative learning rates, meaning that the encoder model is frozen for one epoch while the feed-forward is optimized with a learning rate. After the first epoch, the entire model is fine-tuned with a different learning rate. Since we are fine-tuning on a small dataset, we make use of early stopping with a patience of 3. The best saved checkpoint is decided using the overall Kendall-tau correlation on the test set. The training hyper-parameters used are given in Table~\ref{tab:app_hyper}. Since we have a total of 7000 annotated segments, we perform a 3 fold cross validation split (500 training and 2000 testing) and ensure that the English sentences in the test set are not present during training. We report the mean correlation values for each language. The variance was observed to be less than 0.02. A similar experiment setup was followed for the zero shot evaluation of Indic-COMET, where additionally training segments belonging to a particular language was dropped from the training dataset.

\section{Additional Results}
\label{sec:appendix_add_results}



Table \ref{tab:max_min} shows the maximum and minimum values of each metric on our dataset, across all languages. Note that while some of the metrics are bounded by a theoretical minimum and maximum, some others (especially the trained metrics) are not strictly restricted to a specific scoring range. It would be possible to see a lower minimum value or a higher maximum value on other datasets with such metrics. \\

Figure \ref{intervalviolin} shows metric scores for different human score intervals (0-5, 6-10, 11-20, 21-25). This helps analyse whether the metric scores are roughly in the same buckets or same range as human-scores without focusing on the fine-grained ratings that might not always be of significance. From the plots, we observe that high-performing metrics such as BERTScores and COMET-DA correlate positively with the metric scores as the human scores increase. However, poor-performing metrics on Indic languages such as PRISM (due to lack of training data for Indic languages) do not have correlated metric v/s human spreads even at a coarse-level.\\

Figure \ref{scatter} depicts a scatter plot of metric scores on the y-axis against human scores on the x-axis. The scatter plots provide more insights than just the correlation values \cite{DBLP:conf/acl/MathurBC20_tangled}. We note that the metrics falter by producing some false high and false-low scores.
However, the metrics produce a higher density of decently correlated scores to produce a  net positive correlation trend in most cases.\\




Table \ref{tab:avgcompscore} shows the average scores per system considering the scores provided by the annotator on all the outputs from that system. We find that the best performing model changes across the 5 languages. For Hindi, Malayalam and Tamil, IndicTrans outputs are found to get higher scores on average. For Malayalam Bing API is a close-second and NLLB for Tamil. For Gujarati Bing-API is the best performing, with IndicTrans and NLLB performances being very close. In case of Marathi, NLLB outputs are better, followed by IndicTrans. Averaging further across all 5 languages, IndicTrans is found to be the highest scoring model.\\

Table~\ref{tab:intr_eval_error_type} contains the correlation values for the various metrics on the Fluency-only and Accuracy-only error subsets discussed in section~\ref{sec:corr_flu_ade}. We observe that all the metrics on average correlate better with the human scores when only accuracy errors are annotated compared to having only fluency errors.

\begin{table*}[t]
\small
\centering

\begin{tabular}{lcccccccccc}
\toprule
& \multicolumn{2}{c}{\textbf{gu}}                  
& \multicolumn{2}{c}{\textbf{hi}}     
& \multicolumn{2}{c}{\textbf{mr}}                               
& \multicolumn{2}{c}{\textbf{ml}}                               
& \multicolumn{2}{c}{\textbf{ta}}                               
\\
\multirow{-2}{*}{\bf Metric}                     
& \textbf{Flu} 
& \textbf{Acc}              
& \textbf{Flu} 
& \textbf{Acc}   
& \textbf{Flu} 
& \textbf{Acc}               
& \textbf{Flu} 
& \textbf{Acc}                
& \textbf{Flu} 
& \textbf{Acc}                
\\
\midrule
BLEU-1                               
& 0.138 & 0.268
& 0.067 & 0.151       
& 0.162 & 0.215
& 0.212 & 0.388
& 0.145 & 0.371 
\\
BLEU-2                               
& 0.123   & 0.249              
& 0.074   & 0.155            
& 0.199   & 0.211                        
& 0.192   & 0.348                    
& 0.161   & 0.312
\\
BLEU-3                               
& 0.126 & 0.242                      
& 0.077 & 0.159         
& 0.202 & 0.203                         
& 0.18  & 0.313                       
& 0.162 & 0.275                      
\\
BLEU-4                               
& 0.13           & 0.227                
& 0.078           & 0.156  
& 0.208            & 0.18               
& 0.186          & 0.29           
& 0.158           & 0.28               
\\
SacreBLEU                            
& 0.112           & 0.246                  
& 0.076        & 0.156     
& 0.224              & 0.212            
& 0.194            & 0.338                
& 0.154            & 0.331              
\\
ROUGE-L                              
& 0.126       & 0.247                  
& 0.061        & 0.154     
& 0.182        & 0.196                   
& 0.22        & 0.352                    
& 0.164       & 0.334                   
\\
chrF++                               
& 0.1        & 0.309                   
& 0.047      & 0.164    
& 0.171      & 0.25                   
& 0.169      & 0.413                    
& 0.161     & 0.413            
\\
TER                                  
& 0.127       & 0.232                    
& 0.072       & 0.154        
& 0.18        & 0.209                 
& 0.237        & 0.341                 
& 0.15        & 0.317                  
\\
\midrule
EA 
& 0.076         & 0.19                 
& -0.004        & 0.091   
& 0.135          & 0.171                 
& 0.184         & 0.363                 
& 0.069       & 0.362                   
\\
VE            
& 0.143       & 0.27                   
& 0.052       & 0.172     
& 0.115      & 0.214                     
& 0.217        & 0.356                   
& 0.146       & 0.376                  
\\
GM         
& 0.13       & 0.265                   
& 0.038      & 0.142       
& 0.18       & 0.219                    
& 0.214      & 0.383                   
& 0.187     & 0.42                     
\\
LASER                         
& 0.102         & 0.171                
& -0.056        & 0.099
& 0.111         & 0.186                   
& 0.161         & 0.393              
& 0.011         & 0.189               
\\
LabSE  
& 0.086       & 0.342                  
& -0.064      & 0.116 
& 0.093       & 0.292               
& 0.155       & 0.44                   
& 0.127       & 0.427  
\\
\midrule
mBERT                                
& 0.099        & 0.313                 
& 0.068          & 0.209     
& 0.168       & 0.278                    
& 0.23         & 0.434                  
& 0.159         & 0.435                
\\
distilmBERT                          
& 0.075         & 0.309                 
& 0.063         & 0.196   
& 0.145         & 0.249                 
& 0.226         & 0.42                 
& 0.14          & 0.409                 
\\
IndicBERT                            
& 0.111      & 0.31                   
& 0.063        & 0.184     
& 0.18          & 0.276                     
& 0.217         & 0.425                
& 0.158         & 0.437                     
\\
MuRIL                                
& 0.093      & 0.331                   
& 0.063     & 0.203      
& 0.165      & 0.283                    
& 0.229      & 0.436                     
& 0.18       & 0.461  
\\
\midrule
PRISM                                
& 0.04            & 0.006              
& -0.051          & 0.078  
& 0.078           & 0.133               
& 0.001           & 0.115             
& -0.087          & 0.068              
\\
BLEURT-20                            
& 0.066            & 0.367             
& -0.016           & 0.194
& 0.155            & 0.341             
& 0.232            & 0.451               
& 0.193            & 0.457                   
\\
COMET-DA                             
& 0.174      & 0.412                   
& 0.121      & 0.313      
& 0.167     & 0.38                  
& 0.254    & 0.503                     
& 0.308    & 0.525                     
\\
COMET-MQM
& 0.140  & 0.317
& 0.017    &  0.221  
&  0.130  &    0.298               
& 0.242  &  0.379                     
& 0.240  & 0.466                     
\\ 
\bottomrule
\end{tabular}
\caption{Kendall-tau ($\tau$) correlations of the different metrics with the two MQM subsets (Fluency (Flu.) and Accuracy (Acc.)) across the 5 languages.}
\label{tab:intr_eval_error_type}
\end{table*}

%% file: violinandscatterplots.tex
\begin{figure*}[ht]
    


        \raggedleft
        \includegraphics[scale=0.4]{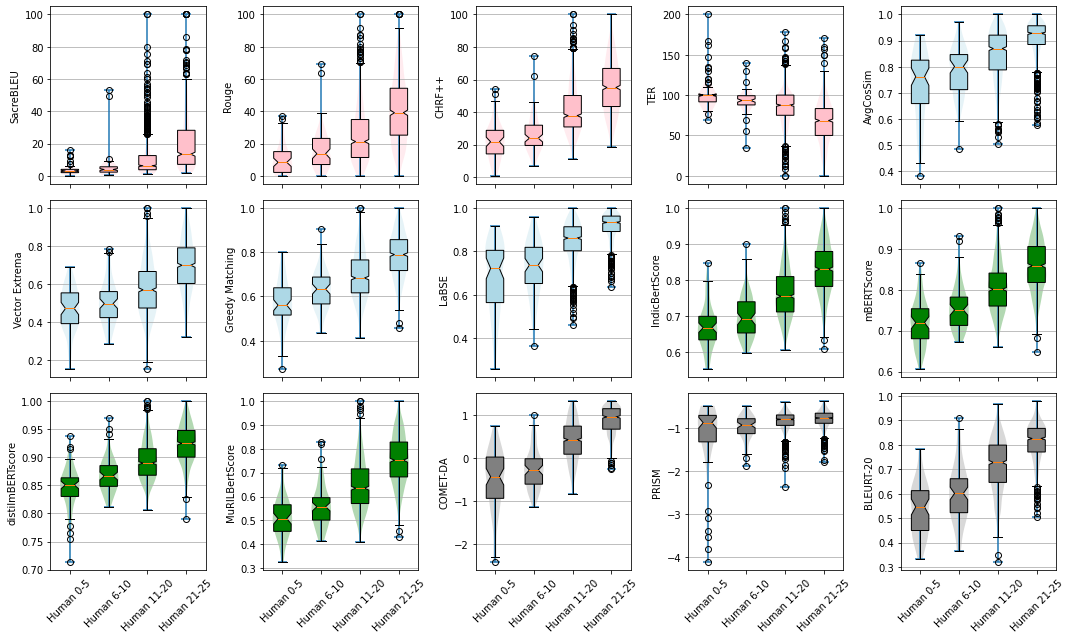}
        \caption{Metric scores for different human score intervals for Malayalam}
        \label{intervalviolin}

\end{figure*}

\begin{figure*}[ht]
        \centering
        \includegraphics[scale=0.55]{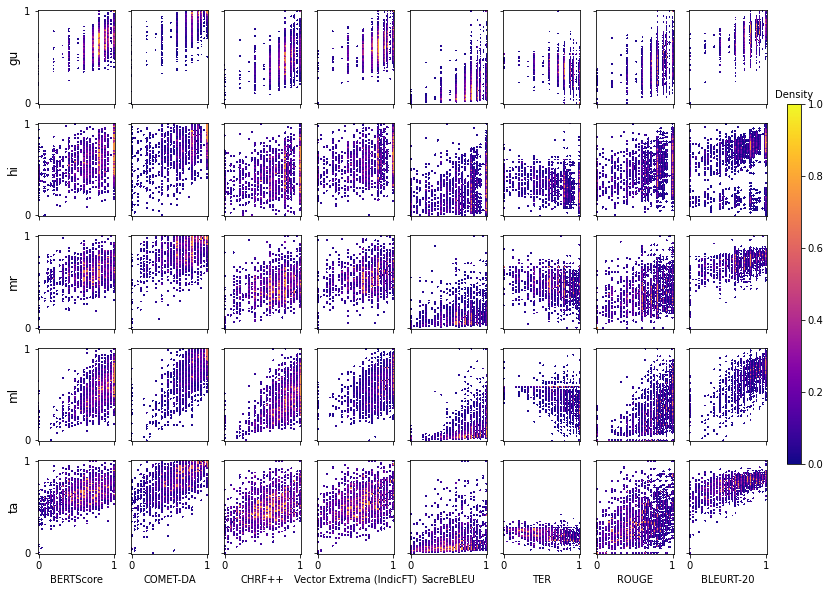}
        \caption{Metric scores vs Human scores. The density colour map is used to indicate whether higher or fewer number of points overlap in the coloured region}
        \label{scatter}

\end{figure*}

